%% file: iclr2025_workshop.tex
\documentclass{article} 
\usepackage{iclr2025_conference,times}

\include{math_commands.tex}

\usepackage{hyperref}
\usepackage{url}
\usepackage{array}  
\usepackage{boldline}
\usepackage{booktabs}  
\usepackage{xcolor}    
\usepackage{colortbl}  
\usepackage{siunitx}   
\usepackage{makecell}  
\usepackage{caption}   
\usepackage{graphicx}

\usepackage{amsmath,amsfonts,amssymb}       %
\usepackage{nicefrac}       %
\usepackage{microtype}      %
\usepackage{float}

\usepackage{algorithm,algorithmic}

\usepackage{wrapfig}

\usepackage{microtype}
\usepackage{graphicx}
\usepackage{subcaption}
\usepackage{chngcntr}
\definecolor{lightgray}{gray}{0.9}
\definecolor{midgray}{gray}{0.7}
\usepackage{enumitem}

\usepackage{amsmath}
\usepackage{amssymb}
\usepackage{mathtools}
\usepackage{amsthm}
\usepackage{textcomp}
\usepackage{newtxmath}
\newcommand{\std}[1]{\textcolor{black}{\scriptsize{$\pm #1$}}}
\newcommand{\highlight}[1]{\cellcolor{blue!10}{#1}}

\title{Shaping Inductive Bias in Diffusion Models through Frequency-Based Noise Control}

\author{Thomas Jiralerspong \\
Mila - Quebec AI Institute and Université de Montréal, Quebec\\
\texttt{thomas.jiralerspong@mila.quebec} \\
\AND
Berton Earnshaw, Jason Hartford \\
Valence Labs \\
\texttt{\{berton.earnshaw,jason.hartford\}@recursion.com} \\
\AND
Yoshua  Bengio, Luca Scimeca \\
Mila - Quebec AI Institute and Université de Montréal, Quebec\\
\texttt{\{yoshua.bengio,luca.scimeca\}@mila.quebec} 
}

\iclrfinalcopy 
\begin{document}

\maketitle

\begin{abstract}
Diffusion Probabilistic Models (DPMs) are powerful generative models that have achieved unparalleled success in a number of generative tasks. In this work, we aim to build inductive biases into the training and sampling of diffusion models to better accommodate the target distribution of the data to model. For topologically structured data, we devise a frequency-based noising operator to purposefully manipulate, and set, these inductive biases. We first show that appropriate manipulations of the noising forward process can lead DPMs to focus on particular aspects of the distribution to learn. We show that different datasets necessitate different inductive biases, and that appropriate frequency-based noise control induces increased generative performance compared to standard diffusion. Finally, we demonstrate the possibility of ignoring information at particular frequencies while learning. We show this in an image corruption and recovery task, where we train a DPM to recover the original target distribution after severe noise corruption.
\end{abstract}


\section{Introduction}
\label{sec:intro}

Diffusion Probabilistic Models (DPMs) have recently emerged as powerful tools for approximating complex data distributions, finding applications across a variety of domains, from image synthesis to probabilistic modeling \citep{yang_diffusion_2024, ho_denoising_2020, sohl-dickstein_deep_2015, venkatraman2024amortizing, sendera2024diffusion}. These models operate by gradually transforming data into noise through a defined diffusion process and training a denoising model~\citep{vincent2008extracting,alain2014regularized} to learn to reverse this process, enabling the generation of samples from the desired distribution via appropriate scheduling. Despite their success, the inductive biases inherent in diffusion models remain largely unexplored, particularly in how these biases influence model performance and the types of distributions that can be effectively modeled.

Inductive biases are known to play a crucial role in deep learning models, guiding the learning process by favoring certain types of data representations over others \citep{geirhos2019imagenet, bietti2019inductive, tishby2015deep}. A well-studied example is the Frequency Principle (F-principle) or spectral bias, which suggests that neural networks tend to learn low-frequency components of data before high-frequency ones \citep{xu_training_2019, rahaman_spectral_2019}. Another related phenomenon is what is also known as the simplicity bias, or shortcut learning \citep{Geirhos2020, scimeca2021which, scimeca2023shortcut}, in which models are observed to preferentially pick up on simple, easy-to-learn, and often spuriously correlated features in the data for prediction. 
If left implicit, it is often unclear whether these biases will improve or hurt the performance of generative model on downstream task, and they could lead to flawed approximations\citep{scimeca2023leveraging}.
 In this work, we aim to explicitly tailor the  
inductive biases of DPMs to better learn the target distribution of interest.

Recent studies have begun to explore the inductive biases inherent in diffusion models. For instance, Kadkhodaie et al. (2023) analyze how the inductive biases of deep neural networks trained for image denoising contribute to the generalization capabilities of diffusion models. They demonstrate that these biases lead to geometry-adaptive harmonic representations, which play a crucial role in the models' ability to generalize beyond the training data~\citep{kadkhodaie2023generalization}.
Similarly, Zhang et al. (2024) investigate the role of inductive and primacy biases in diffusion models, particularly in the context of reward optimization. They propose methods to mitigate overoptimization by aligning the models' inductive biases with desired outcomes~\citep{zhang2024confronting}. Other methods, such as noise schedule adaptations~\citep{sahoo_diffusion_2024} and the introduction of non-Gaussian noise \citep{bansal_cold_2022} have shown promise in improving the performance of diffusion models on various tasks. However, the exploration of frequency domain techniques within diffusion models is a relatively new area of interest. One of the pioneering studies in this domain investigates the application of diffusion models to time series data, where frequency domain methods have shown potential for capturing temporal dependencies more effectively \citep{crabbé2024timeseriesdiffusionfrequency}. Similarly, the integration of spatial frequency components into the denoising process has been explored for enhancing image generation tasks \citep{Qian_2024_CVPR, yuan2023spatialfrequencyunetdenoisingdiffusion}, showcasing the importance of considering frequency-based techniques as a means of refining the inductive biases of diffusion models.

In this work, we explore a new avenue, to build inductive biases in DPMs by frequency-based noise control. The main hypothesis in this paper is that the noising operator in a diffusion model has a direct influence on the model's representation of the data. Intuitively, the information erased by the noising process is the very information that the denoising model has pressure to learn, so that reconstruction is possible. Accordingly, we propose that by strategically manipulating the noising operation, we can effectively steer the model to learn particular aspects of the data distribution. We focus our attention to the generative learning of topologically structured data, and propose an approach that involves designing a frequency-based noise schedule that selectively emphasizes or de-emphasizes certain frequency components during the noising process. In this paper, we refer to our approach as \emph{frequency diffusion}. Because the Fourier transform of a Gaussian is just another Gaussian in the frequency domain, this approach allows us to maintain the Gaussian assumptions of the diffusion process while reorienting the noising operator within the frequency domain, enabling the generation of Gaussian noise at different frequencies and thereby influencing the model’s learning trajectory.

We report several findings. First, we show that when the information content in the data lies more heavily in particular frequencies, frequency diffusion yields better samplers. Furthermore, we test this in several natural datasets, and show that depending on the dataset characteristic, different settings of our frequency diffusion approach yield optimal results, often with comparable or superior performance to standard diffusion. Finally, we show that through frequency-denoising we can recover complex distributions after severe noise corruption at particular frequencies, opening interesting venues for applications within the generative landscape. 

We summarize our contributions as follows:

\begin{enumerate}
\item We introduce a frequency-informed noising operator that can shape the inductive biases of diffusion models.
\item We empirically show that \emph{frequency diffusion} can steer models to better approximate information at particular frequencies of the underlying data distribution.
\item We provide empirical evidence that models trained with frequency-based noise schedules can outperform traditional diffusion schedules across multiple datasets.
\item We show that through frequency-denoising we can recover complex distributions after severe noise corruption at particular frequencies.
\end{enumerate}


\label{sec:related_work}







\section{Methods} \label{sec:methods}

\subsection{Denoising Probabilistic Models (DPMs)}\label{sec:methods:dpms}

Denoising Probabilistic Models, are a class of generative models that learn to reconstruct complex data distributions by reversing a gradual noising process. DPMs are characterized by a \emph{forward} and \emph{backward} process. The \emph{forward process} defines how data is corrupted, typically by Gaussian noise, over time. Given a data point $\mathbf{x}_0$ sampled from the data distribution $q(\mathbf{x}_0)$, the noisy versions of the data $\mathbf{x}_1, \mathbf{x}_2, \ldots, \mathbf{x}_T$ are generated according to:

\begin{equation}
q(\mathbf{x}_t \mid \mathbf{x}_{t-1}) = \mathcal{N}(\mathbf{x}_t; \sqrt{\alpha_t} \mathbf{x}_{t-1}, (1 - \alpha_t)\mathbf{I})
\end{equation}

with $\alpha_t$ variance schedule.  The \emph{reverse process} models the denoising operation, attempting to recover $\mathbf{x}_{t-1}$ from $\mathbf{x}_t$:

\begin{equation}
p_\theta(\mathbf{x}_{t-1} \mid \mathbf{x}_t) = \mathcal{N}(\mathbf{x}_{t-1}; \mu_\theta(\mathbf{x}_t, t), \sigma^2_t \mathbf{I}),
\end{equation}

where $\mu_\theta(\mathbf{x}_t, t)$ is predicted by a neural network $f_\theta$, and the variance $\sigma^2_t$ is can be fixed, learned, or precomputed based on a schedule. We often train the denoising model by minimizing a variational bound on the negative log-likelihood:

\begin{equation}
L = \mathbb{E}_{t, \mathbf{x}_0, \mathbf{\epsilon}} \left[ \left\| \mathbf{\epsilon} - \mathbf{\epsilon}_\theta(\mathbf{x}_t, t) \right\|^2 \right]
\end{equation}

where $\mathbf{\epsilon}$ is the Gaussian noise added to $\mathbf{x}_0$, and $\mathbf{\epsilon}_\theta$ is the model’s prediction of this noise. To generate new samples, we sample from a Gaussian distribution and apply the learned reverse process iteratively, often starting from a sample drawn from a simple Gaussian noise distribution.

\subsection{Frequency Diffusion}\label{sec:methods:freq_noise}

\begin{figure*}
    \centering
    \includegraphics[width=\textwidth]{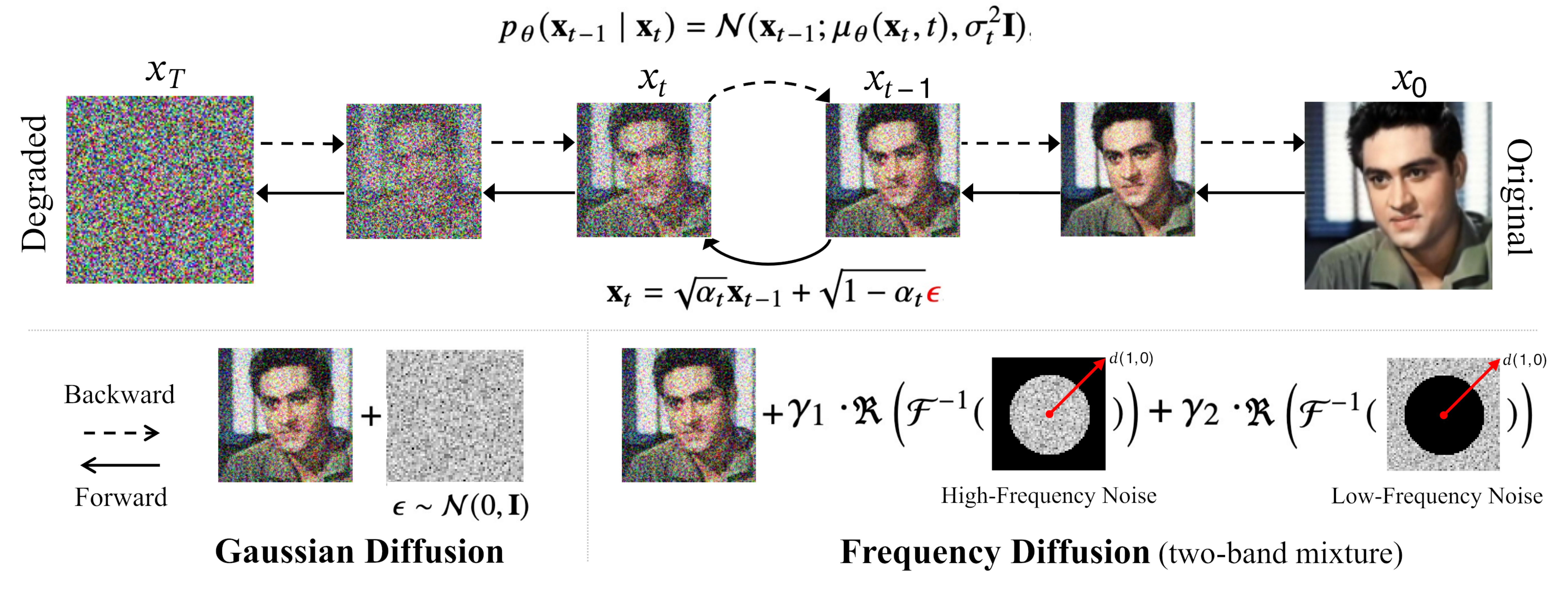}
    \caption{Frequency diffusion under a generalized framework.}
    \label{fig:freq_diffusion}
\end{figure*}

The objective of this section is to generate spatial Gaussian noise whose frequency content can be systematically manipulated according to an arbitrary weighting function. In \autoref{sec:methods:dpms}, we describe how $ \mathbf{x}_t $ is obtained from $ \mathbf{x}_{t-1} $ by adding Gaussian noise sampled from a normal distribution to the sample at time step $ t-1 $. Specifically, we can sample $ \mathbf{\epsilon}_t \sim \mathcal{N}(0, \mathbf{I}) $ and obtain $\mathbf{x}_t$ as:
\begin{equation}
\mathbf{x}_t = \sqrt{\alpha_t}\,\mathbf{x}_{t-1} \;+\; \sqrt{1 - \alpha_t}\,\mathbf{\epsilon}.
\end{equation}

Let us denote by \(\mathbf{x}\in\mathbb{R}^{H\times W}\) an image (or noise field) in the spatial domain, and by \(\mathcal{F}\) the two-dimensional Fourier transform operator. We let \(\mathbf{N}_\text{freq} \in \mathbb{C}^{H\times W}\) be a complex-valued random field whose real and imaginary parts are i.i.d.~Gaussian:
\begin{equation}
\mathbf{N}_\text{freq} 
\;=\; 
\mathbf{N}_\text{real} \;+\; i\,\mathbf{N}_\text{imag},
\quad
\mathbf{N}_\text{real}, \,\mathbf{N}_\text{imag} 
\;\sim\; 
\mathcal{N}\bigl(0, \mathbf{I}\bigr).
\end{equation}
where each pixel (or frequency bin) in \(\mathbf{N}_\text{real}\) and \(\mathbf{N}_\text{imag}\) is drawn independently from a standard normal distribution. We introduce a \emph{weighting function} \(w(f_x, f_y)\) that scales the amplitude of each frequency component.  Let \(\mathbf{f} = (f_x, f_y)\) denote coordinates in frequency space, where $f_x \;=\; \frac{k_x}{W}$, $f_y \;=\; \frac{k_y}{H}$, and $k_x, k_y$ are integer indices (ranging over the width and height), while \(H\) and \(W\) are the image dimensions. We define the frequency-controlled noise $\mathbf{N}_\text{freq}^{(w)}(\mathbf{f})$ as:
\begin{equation} \label{eq:general_weight}
\mathbf{N}_\text{freq}^{(w)}(\mathbf{f})
\;=\;
\mathbf{N}_\text{freq}(\mathbf{f})
\;\odot\;
w(\mathbf{f}),
\end{equation}
After applying \(w(\mathbf{f})\) in the frequency domain, we invert back to the spatial domain to obtain \(\mathbf{\epsilon}^{(w)}\), our \emph{frequency-shaped} noise:
\begin{equation} \label{eq:general_noise_spatial}
\mathbf{\epsilon}^{(w)} 
\;=\; 
\Re\Bigl(\mathcal{F}^{-1}\bigl(\mathbf{N}_\text{freq}^{(w)}\bigr)\Bigr),
\end{equation}
where \(\Re(\cdot)\) denotes the real part, ensuring that our final noise field is purely real. 

In summary, any frequency weighting can be represented in this unified framework:
\begin{equation*}
\mathbf{\epsilon}
\xrightarrow{\;\mathcal{F}\;}
\mathbf{N}_\text{freq}
\xrightarrow{\;w(\mathbf{f})\;}
\mathbf{N}_\text{freq}^{(w)}
\xrightarrow{\;\mathcal{F}^{-1}\;}
\mathbf{\epsilon}^{(w)}.
\end{equation*}
With this, we have a simple mechanism for generating noise whose power spectrum can purposefully controlled. Note that standard white Gaussian noise is a special case of this formulation, where \(w(\mathbf{f})=1\) for all \(\mathbf{f}\). In contrast, more sophisticated weightings allow one to emphasize, de-emphasize, or even remove specific bands of the frequency domain.

\subsection{Frequency Noise operators}
\label{supp:sec:freq_noise_operator}

In this work, the design of $w(\mathbf{f})$ is especially important. In this section, we propose several alternatives, while showing empirical results on a particular choice of $w(\mathbf{f})$. 

\subsubsection*{Power-Law Weighting.}
A natural alternative choice is the power-law weighting, expressed as:
\begin{equation}
w(\mathbf{f}) \;=\; \|\mathbf{f}\|^{\alpha},
\end{equation}
where \(\mathbf{f} = (f_x,f_y)\) denotes a frequency coordinate, and the exponent \(\alpha\) determines which frequencies are amplified or suppressed.
Power-law weighting is popular in the modeling if natural phenomena (e.g., fractal landscapes, turbulence) where the energy distribution often follows an approximate power spectrum \citep{van1996modelling}.

\subsubsection*{Exponential Decay Weighting}
Another alternative is an exponential decay function, defined as as:
\begin{equation}
w(\mathbf{f}) \;=\; \exp\!\bigl(-\beta\,\|\mathbf{f}\|^{2}\bigr),
\end{equation}
where \(\beta>0\), and frequencies with larger norms \(\|\mathbf{f}\|\) are exponentially suppressed. This weighting effectively imposes spatial correlations, e.g. for \(\beta\) close to 0 the function induces the retention of more high-frequency components, while for large \(\beta\), the function quickly damps out high frequencies, resulting in a smoothing of the spatial domain.

\subsubsection*{Band-Pass Masking and Two-Band Mixture}
Finally, a \emph{band-pass mask} can be viewed as a special case of a more general weighting function:
\begin{equation}
w(\mathbf{f}) \;\in\; \{0,1\}.
\end{equation}
In this case, the frequency domain is split into a set of permitted and excluded regions, or radial thresholds. We this, we can construct several types of filters, including a low-pass filter retaining only frequencies below a cutoff (e.g., \(\|\mathbf{f}\| \leq \omega_c\)) a high-pass filter keeping only frequencies above a cutoff, or more generally a filter restricting \(\|\mathbf{f}\|\) to lie between two thresholds \([a_{\mathrm{min}},b_{\mathrm{max}}]\).
We thus define a simple band pass filter as:
\begin{equation}
w(\mathbf{f}) \;=\; 
\mathbf{M}_{[a,b]}(f_x, f_y)
\;=\;
\begin{cases}
1, & \text{if } a \leq d(f_x, f_y) \leq b, \\
0, & \text{otherwise}.
\end{cases}
\end{equation}
Here, $d(f_x, f_y) = \sqrt{\left(f_x - \tfrac{1}{2}\right)^2 + \left(f_y - \tfrac{1}{2}\right)^2}$ measures the radial distance in frequency space. In this special case, $ w(\mathbf{f}) $ is simply a \emph{binary} mask, selecting only those frequencies within $ [a, b] $. 

For the experiments in this paper we formulate a simple two-band mixture, where, we limit ourselves to constructing noise as a simple linear combination of two band-pass filtered noise components. Specifically, as in the original band-based approach, we generate frequency-filtered noise $\mathbf{\epsilon}_f$ via:
\begin{equation} \label{eq:high_low_formulation}
\mathbf{\epsilon}_f 
\;=\; 
\gamma_l \,\mathbf{\epsilon}_{[a_l, b_l]}
\;+\;
\gamma_h \,\mathbf{\epsilon}_{[a_h, b_h]},
\end{equation}
where $ \gamma_l $ and $ \gamma_h $ denote the relative contributions of a low- and a high-frequency noise components, each filtering noise respectively in the ranges $[a_l, b_l]$ (low-frequency range) and $[a_h, b_h]$ (high-frequency range).
We uniquely refer to $\epsilon_{[a, b]}$ as the noise filtered in the $[a, b]$ frequency range following \autoref{eq:general_weight} and \autoref{eq:general_noise_spatial}. Standard Gaussian noise emerges as a particular instance (with $ \gamma_l = 0.5 $, $ \gamma_h = 0.5 $, $ a_l = 0 $, $ b_l = 0.5 $, $ a_h = 0.5 $, and $ b_h = 1 $) of this formulation.

\subsection{datasets} \label{suppl:sec:datasets}
For the experiments, we consider five datasets, namely: MNIST, CIFAR-10, Domainnet-Quickdraw, Wiki-Art and CelebA; providing examples of widely different visual distributions, scales, and domain-specific statistics.

\paragraph{MNIST:} MNIST consists of \(70,000\) grayscale images of handwritten digits (0-9)~\citep{dsprites17}. MNIST provides a simple test-bed to for the hypothesis in this work, as a well understood dataset with well structured, and visually coherent samples.

\paragraph{CIFAR-10:} CIFAR-10 contains \(60,000\) color images distributed across 10 object categories \citep{krizhevsky2009learning}. The dataset is highly diverse in terms of object appearance, backgrounds, and colors, with the wide-ranging visual variations across classes like animals, vehicles, and other common objects.

\paragraph{DomainNet-Quickdraw:} DomainNet-Quickdraw features \(120,750\) sketch-style images,
 These images, drawn in a minimalistic, abstract style, present a distribution that is drastically different from natural images, with sparse details and heavy visual simplifications.

\paragraph{WikiArt:} WikiArt consists of over \(81,000\) images of artwork spanning a wide array of artistic styles, genres, and historical periods \citep{saleh2015large}. The dataset encompasses a rich and varied distribution of textures, color palettes, and compositions, making it a challenging benchmark for generative models, which must capture both the global structure and fine-grained stylistic variations that exist across different forms of visual art.

\paragraph{CelebA:} CelebA contains \(202,599\) images of celebrity faces, each \(178 \times 218\) pixels in resolution ~\citep{liu2015faceattributes}. The dataset presents a diverse distribution of human faces with variations in pose, lighting, and facial expressions. 

\section{Results} \label{sec:results}
All experiments involve separately training and testing DPMs with various \emph{frequency diffusion} schedules, as well as baseline standard denoising diffusion training. We use DDPM fast sampling \citep{ho2020denoising} to efficiently generate samples for all reported metrics. Across the experiments, we report FID and KID scores as similarity score estimate metrics of the generated samples with respect to a held-out set of data samples. In all relevant experiments, we compute the metrics on embeddings from block 768 of a pre-trained Inception v3 model.

\subsection{Improved Diffusion Sampling via Frequency-Based Noise Control}
In the first set of experiments, we wish to test our main hypothesis, i.e. that appropriate manipulation of the frequency components of the noise can better support the learning of the distribution of interest. 
\begin{wrapfigure}[26]{r}{0.50\textwidth}
\centering
\vspace*{-.5em}
\includegraphics[width=\linewidth]{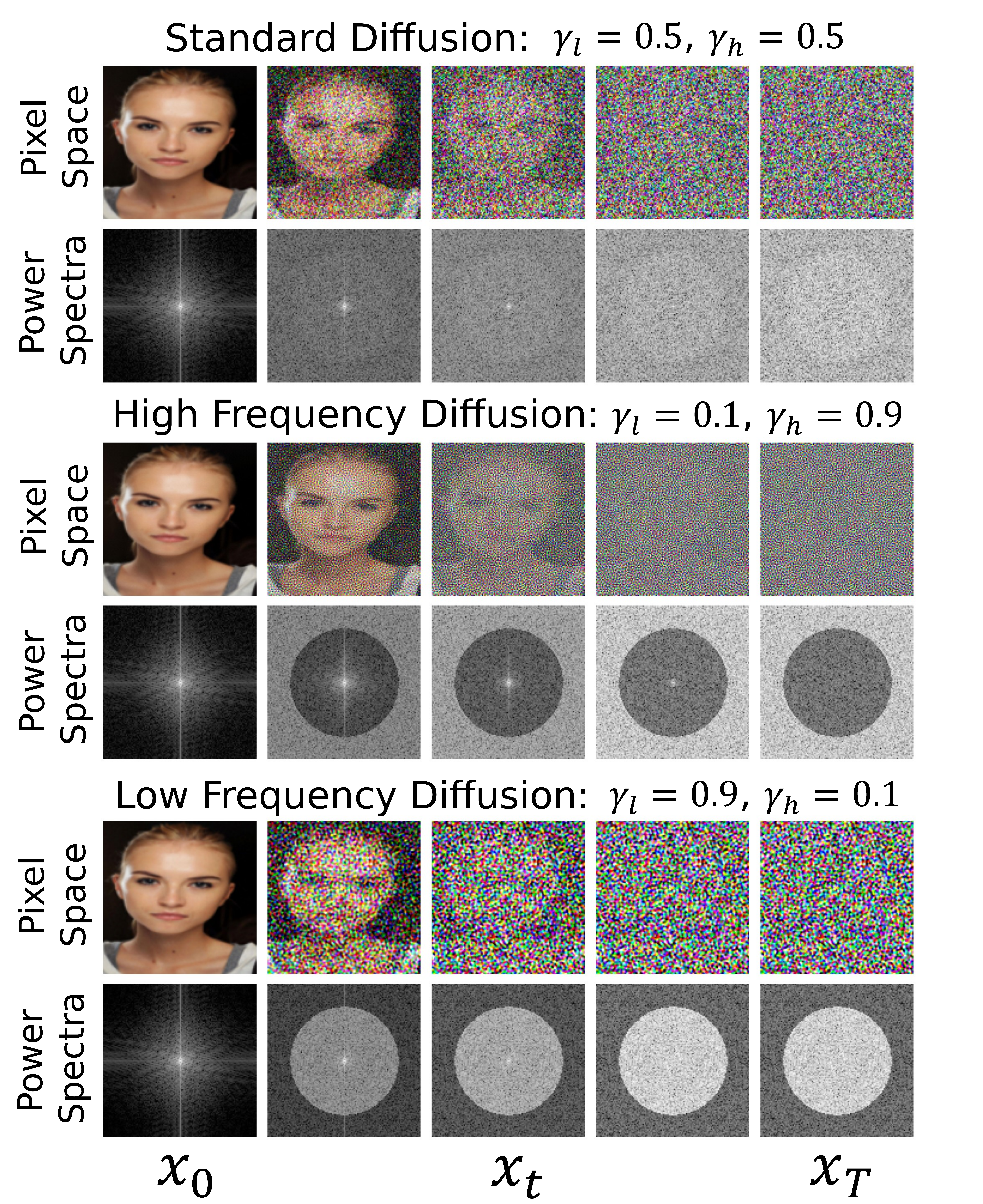}
\caption{Power spectra and image visuals of the forward Process in standard diffusion, as compared to high and low-frequency noise settings of a two-band mixture noise parametrization.} 
\label{fig:frequency_diffusion_noising}
\end{wrapfigure} covering \(345\) object categories \citep{peng2019moment}.
We follow the formulation in \autoref{eq:high_low_formulation} to train and compare diffusion models with a noisy operator prioritizing different parts of the frequency distribution. In these experiments we fix $a_l = 0 $, $ b_2 = 1 $, and $ b_l = a_h = 0.5 $, while performing a linear sweep of the  $\gamma_l$ and $\gamma_h$ parameters by searching $\gamma_l \in [.1, .2, ..., .9]$ and $\gamma_h = 1-\gamma_l$.


\subsubsection{Qualitative Overview}

First, we show a qualitative example of a standard linear noising schedule forward operation in \autoref{fig:frequency_diffusion_noising}, as compared to two particular settings of our constant high and low-frequency linear schedules of the band-pass filter. With standard noise, information is uniformly removed from the image, with sample quality degrading evenly over time. In the high-frequency noising schedule, sharpness and texture are removed more prominently, while in the low-frequency noising schedule, general shapes and homogeneous pixel clusters are affected most, yielding qualitatively different information destruction operations. As discussed previously, we hypothesize that this will in turn purposely affect the statistics of the information learned by the denoiser model, effectively focusing the diffusion sampling process on different parts of the distribution. 

\begin{wrapfigure}[17]{r}{0.50\textwidth}
\includegraphics[width=.9\linewidth]{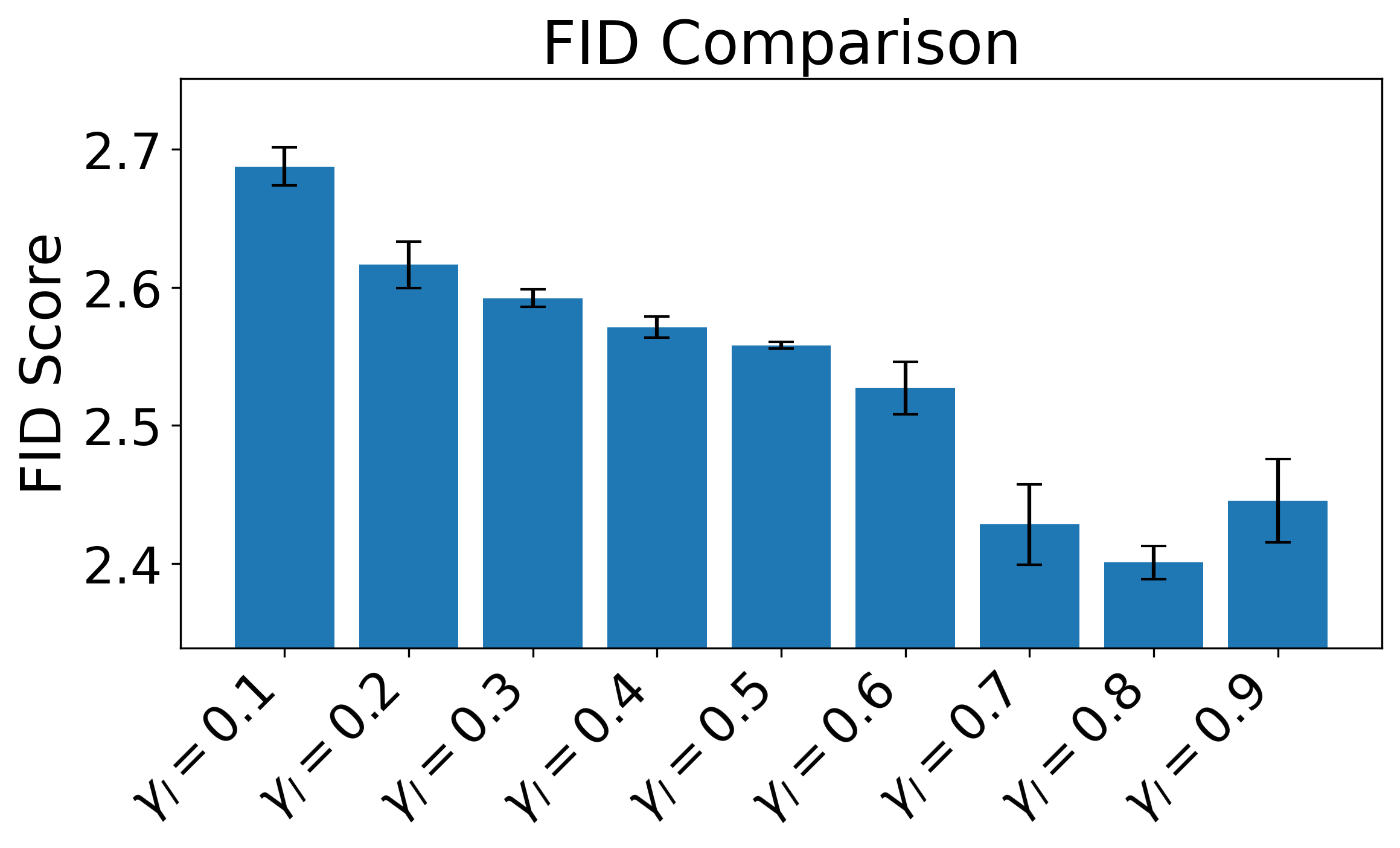}
\caption{FID of diffusion samplers trained with various combinations of frequency noise. The settings for $\gamma_l=0.5$ yields standard diffusion training.}
\label{fig:corrupt_noise_learning}
\end{wrapfigure}
\subsubsection{Learning Target Distributions from Frequency-Bounded Information}
We conduct experiments to learn the distribution of data where, by construction, the information content lies in the low frequencies. We use the CIFAR-10 dataset, and corrupt the original data with high-frequency noise $\mathbf{\epsilon}_{[.3, 1.]}$, thus erasing the high-frequency content while predominantly preserving the low-frequency details in the range $\mathbf{\epsilon}_{[0., .3]}$. We train 9 diffusion models, including a standard diffusion (\emph{baseline}) model, and 8 models trained with frequency-based noise control spanning 8 combinations of $\gamma_l$ ($\gamma_h=1-\gamma_l$). We repeat the experiment over three seeds and report the average FID and error in \autoref{fig:corrupt_noise_learning}. In the figure, we observe the DPMs trained with higher amounts of low-frequency noise (higher $\gamma_l$) to perform significantly better than both the baseline ($\gamma_l=0.5$), and higher frequency denoising models (lower $\gamma_l$). Furthermore, we see a mostly monotonically descending trend in FID for increasing values of lower frequency noise in the diffusion forward schedule, supporting the original intuition of how the frequency manipulation of the noising operator can directly steer the denoiser's learning trends, and therefor how progressively higher amount of low-frequency forward noise aid in the learning of samplers for data containing mostly low-frequency information.

\subsubsection{Frequency-Based noise control in natural datasets} \label{sec:qualitative_freq_results}

We further test our hypothesis by training 9 models for each of the datasets considered, inclusive of all $\gamma$-variations of our two-band mixture frequency-based noise schedule. We train these models on MNIST, CIFAR-10, Domainnet-Quickdraw, Wiki-Art and CelebA, and report the FID and KID metrics for all ablations in \autoref{tab:freq_qualitative}. In the table, we observe three out of five datasets to significantly benefit from frequency-controlled noising schedules, achieving the lowest FID and KID scores across all tested models. Interestingly, the performance trends are also mostly monotonic, which together with our previous experiments is indicative of where the learned information lies. For simple datasets, such as MNIST or CIFAR-10, most frequency denoising settings perform well, with balanced high-to-low-frequency schedules performing best overall. Denoisers for Domainnet-Quickdraw and CelebA yield better performance for slightly higher frequency noising schedules, suggesting higher frequency information content for good FID and KID approximations, while Wiki-Art shows slight biases towards lower frequency schedules.

\begin{table}
\centering
\caption{Results for FID and KID across different settings of $(\gamma_l, \gamma_h)$ for our frequency diffusion two-band mixture schedule across different datasets  (mean $\pm$ standard error across 3 seeds).  The baseline runs correspond to $\gamma_l=\gamma_h=0.5$. }
\label{tab:freq_qualitative}
\resizebox{1\linewidth}{!}{
\begin{tabular}{@{}lcccccccccc@{}}
\toprule
\textbf{Dataset} $\rightarrow$ & \multicolumn{2}{c}{\textbf{MNIST}} & \multicolumn{2}{c}{\textbf{CIFAR-10}} & \multicolumn{2}{c}{\textbf{Domainnet-Quickdraw}} & \multicolumn{2}{c}{\textbf{Wiki-Art}} & \multicolumn{2}{c}{\textbf{CelebA}} \\
\cmidrule(lr){2-3}\cmidrule(lr){4-5}\cmidrule(lr){6-7}\cmidrule(lr){8-9}\cmidrule(lr){10-11}
\textbf{Algo} $\downarrow$ \textbf{Metric} $\rightarrow$ & FID ($\downarrow$) & KID ($\downarrow$) & FID ($\downarrow$) & KID ($\downarrow$) & FID ($\downarrow$) & KID ($\downarrow$) & FID ($\downarrow$) & KID ($\downarrow$) & FID ($\downarrow$) & KID ($\downarrow$) \\
\midrule
$\ \ \ \ \ $baseline & \highlight{0.0168}\std{0.0010} & \highlight{0.0000}\std{0.0000} & \highlight{0.1055}\std{0.0042} & \highlight{0.0001}\std{0.0000} & 0.0875\std{0.0060} & 1.69e-04\std{1.61e-05} & 0.1622\std{0.0133} & 2.53e-04\std{1.80e-05} & 0.0863\std{0.0094} & \highlight{0.0001}\std{0.0000} \\
$\gamma_l=0.1, \gamma_h=0.9$ & 0.2624\std{0.2184} & 7.90e-04\std{6.85e-04} & 0.2648\std{0.0691} & 4.31e-04\std{1.30e-04} & 0.5250\std{0.3907} & 1.46e-03\std{1.21e-03} & 0.2673\std{0.0273} & 4.31e-04\std{4.56e-05} & 0.1555\std{0.0273} & 2.97e-04\std{6.93e-05} \\
$\gamma_l=0.2, \gamma_h=0.8$ & 0.0432\std{0.0187} & 1.10e-04\std{5.24e-05} & 0.2191\std{0.0223} & 3.86e-04\std{6.72e-05} & 0.1843\std{0.0723} & 4.20e-04\std{2.15e-04} & 0.2048\std{0.0063} & 3.43e-04\std{1.27e-05} & 0.1024\std{0.0045} & 1.85e-04\std{2.72e-06} \\
$\gamma_l=0.3, \gamma_h=0.7$ & 0.0267\std{0.0029} & 6.40e-05\std{8.63e-06} & 0.1506\std{0.0168} & 2.28e-04\std{3.34e-05} & 0.1248\std{0.0375} & 2.70e-04\std{1.13e-04} & 0.1865\std{0.0181} & 2.86e-04\std{2.46e-05} & \highlight{0.0838}\std{0.0107} & 1.44e-04\std{1.89e-05} \\
$\gamma_l=0.4, \gamma_h=0.6$ & 0.0224\std{0.0032} & 5.29e-05\std{8.15e-06} & 0.1131\std{0.0079} & 1.64e-04\std{2.15e-05} & \highlight{0.0799}\std{0.0166} & \highlight{0.0001}\std{0.0000} & 0.1597\std{0.0122} & 2.62e-04\std{3.23e-05} & 0.0875\std{0.0020} & 1.49e-04\std{1.71e-06} \\
$\gamma_l=0.6, \gamma_h=0.4$ & 0.0253\std{0.0039} & 5.81e-05\std{7.63e-06} & 0.1131\std{0.0074} & 1.56e-04\std{1.95e-05} & 0.1128\std{0.0174} & 2.57e-04\std{5.56e-05} & \highlight{0.1348}\std{0.0126} & \highlight{0.0002}\std{0.0000} & 0.1068\std{0.0039} & 2.04e-04\std{1.07e-05} \\
$\gamma_l=0.7, \gamma_h=0.3$ & 0.0363\std{0.0075} & 9.14e-05\std{2.04e-05} & 0.1432\std{0.0203} & 2.19e-04\std{3.66e-05} & 0.1353\std{0.0223} & 2.91e-04\std{6.08e-05} & 0.1561\std{0.0123} & 2.32e-04\std{2.46e-05} & 0.0990\std{0.0082} & 1.84e-04\std{2.12e-05} \\
$\gamma_l=0.8, \gamma_h=0.2$ & 0.0512\std{0.0119} & 1.36e-04\std{3.60e-05} & 0.1898\std{0.0095} & 2.88e-04\std{1.85e-05} & 0.2288\std{0.0737} & 5.85e-04\std{2.21e-04} & 0.2256\std{0.0096} & 3.86e-04\std{3.08e-05} & 0.1053\std{0.0185} & 1.95e-04\std{4.34e-05} \\
$\gamma_l=0.9, \gamma_h=0.1$ & 0.3403\std{0.1513} & 9.74e-04\std{4.47e-04} & 0.3226\std{0.0660} & 5.31e-04\std{1.20e-04} & 0.9827\std{0.4229} & 2.84e-03\std{1.29e-03} & 0.3250\std{0.0270} & 5.57e-04\std{3.22e-05} & 0.2291\std{0.0605} & 4.86e-04\std{1.52e-04} \\
\bottomrule
\end{tabular}
}
\end{table}

\subsection{Selective Learning: Frequency-Based Noise Control to Omit Targeted Information}
Following our original intuition, a denoising model has pressure to learn the very information that is erased by the forward noising operator to achieve successful reconstruction. Conversely, when the noising operator is crafted to leave parts of the original distribution intact, no such pressure exists, and the denoising model can effectively discard the left-out statistics during generation. 

In this section, we perform experiments whereby the original data is corrupted with noise at different frequency ranges. The objective is to manipulate the inductive biases of diffusion denoisers to avoid learning the corruption noise, while correctly approximating the relevant information in the data.  We formulate our corruption process as $\rvx'=A_c(\rvx)$, where: 

\begin{equation}
    A_c(\rvx) =  \rvx + \gamma_c \mathbf{\epsilon}_{f[a_c, b_c]}
\end{equation}

\begin{figure}
    \centering
    \includegraphics[width=1\textwidth]{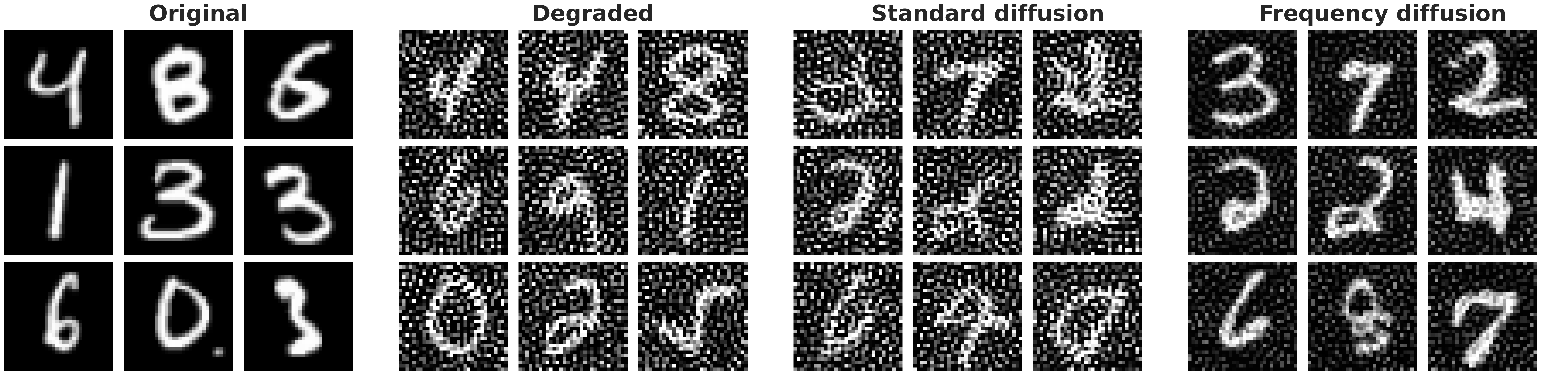}
    \caption{Samples from the original data distribution, the degraded data distribution, a standard diffusion sampler trained on the degraded data distribution, and a \emph{frequency diffusion} sampler trained on the degraded data distribution. We generate noise for data corruption in the frequency range [$a_c=0.5$, $b_c=0.6)$].}
    \label{fig:noise_removal}
\end{figure}

Here, $\mathbf{\epsilon}_{[a_c, b_c]}$ denotes noise in the $[a_c, b_c]$ frequency range. We default $\gamma_c=1.$ and show samples of the original and corrupted distributions in \autoref{fig:noise_removal}. For any standard DPM training procedure, the denoiser would make no distinction of which information to learn, and thus would approximate the corrupted distribution presented at training time. As such, the recovery of the original, noiseless, distribution would normally be impossible. Assuming knowledge of the corruption process, we frame the frequency diffusion learning procedures as a noiseless distribution recovery process, and set $a_l = 0 $, $ b_h = 1 $, $b_l = a_c$, and $a_h = b_c$. This formulation effectively allows for the forward frequency noising operator to omit the range of frequencies in which the noise lies. In line with our previous rationale, this would effectively put no pressure on the denoiser to learn the noise part of the distribution at hand, and focus instead on the frequency ranges where the true information lies.

We compare original and corrupted samples from MNIST, as well as samples from standard and frequency diffusion-trained models in \autoref{fig:noise_removal}. In line with our hypothesis, we observe frequency diffusion DPMs trained with an appropriate frequency noise operator to be able to discard the corrupting information and recover the original distribution after severe noisy corruption. We further measure the FID and KID of the samples generated by the baseline and frequency DPMs against the original (uncorrupted) data samples in \autoref{tab:corruption_fids}. 
We perform 8 ablation studies, considering noises at $0.1$ non-overlapping intervals in the $[0.1, .9]$ frequency range. We observe \emph{frequency diffusion} to outperform standard diffusion training across all tested ranges. Interestingly, we observe better performance (lower FID) for data corruption in the high-frequency ranges, and reduced performance for data corruptions in low-frequency ranges, suggesting a marginally higher information content in the low frequencies for the MNIST dataset.

\begin{table} \label{tab:corruption}
\centering
\caption{Resulting FID and KID between standard diffusion and frequency diffusion DPMs trained on noise-corrupted data, with respect to samples from the true uncorrupted distribution (mean $\pm$ standard error across 3 seeds). We report eight ablation experiments across different non-overlapping corruption noise schemes.}
\label{tab:corruption_fids}
\resizebox{.8\linewidth}{!}{
\begin{tabular}{@{}lcccc@{}}
\toprule
\textbf{Dataset} $\rightarrow$ & \multicolumn{2}{c}{\textbf{Baseline}} & \multicolumn{2}{c}{\textbf{Ours}} \\
\cmidrule(lr){2-3}\cmidrule(lr){4-5}
\textbf{Corruption} $\downarrow$ & FID ($\downarrow$) & KID ($\downarrow$) & FID ($\downarrow$) & KID ($\downarrow$) \\
\midrule
$\epsilon_{[0.1,0.2]}$ & 3.2273\std{8.50e-03} & 0.0114\std{3.13e-05} & \highlight{2.7572\std{3.56e-02}} & \highlight{0.0095\std{1.47e-04}} \\
$\epsilon_{[0.2,0.3]}$ & 3.6601\std{4.43e-03} & 0.0132\std{1.67e-05} & \highlight{3.0416\std{4.47e-02}} & \highlight{0.0107\std{1.79e-04}} \\
$\epsilon_{[0.3,0.4]}$ & 3.4771\std{4.79e-03} & 0.0125\std{1.89e-05} & \highlight{2.9952\std{3.35e-02}} & \highlight{0.0106\std{1.23e-04}} \\
$\epsilon_{[0.4,0.5]}$ & 3.4281\std{5.46e-03} & 0.0123\std{1.98e-05} & \highlight{2.9218\std{2.54e-02}} & \highlight{0.0105\std{8.79e-05}} \\
$\epsilon_{[0.5,0.6]}$ & 3.3638\std{6.31e-03} & 0.0121\std{2.32e-05} & \highlight{2.8267\std{2.81e-02}} & \highlight{0.0102\std{9.32e-05}} \\
$\epsilon_{[0.6,0.7]}$ & 3.2444\std{7.10e-03} & 0.0116\std{2.55e-05} & \highlight{2.7026\std{3.90e-02}} & \highlight{0.0097\std{1.28e-04}} \\
$\epsilon_{[0.7,0.8]}$ & 3.0442\std{6.32e-03} & 0.0109\std{2.29e-05} & \highlight{2.5469\std{6.39e-02}} & \highlight{0.0091\std{2.00e-04}} \\
$\epsilon_{[0.8,0.9]}$ & 3.4660\std{7.90e-03} & 0.0124\std{2.96e-05} & \highlight{2.5138\std{9.63e-02}} & \highlight{0.0090\std{3.07e-04}} \\
\bottomrule
\end{tabular}
}
\end{table}

\section{Discussion and Conclusion} \label{sec:discussion_and_conclusion}

In this work, we studied the potential to build inductive biases in the training and sampling of Diffusion Probabilistic Models by purposeful manipulation of the forward, noising, process. We introduced \emph{frequency diffusion}, an approach that enables us to guide DPMs toward learning specific statistics of the data distribution. We compare \emph{frequency diffusion} to DPS trained with standard gaussian noise on generative visual tasks set by several datasets, with significant varying structure and scales. We show several key findings. First, we show that appropriate manipulation of the forward noising process can serve as a stong inductive bias for diffusion models to better learn the information of the distribution at particular frequencies. Second, we show that this important characteristic can be readily used when training diffusion models on natural dataset, some of which may be better supported by appropriate frequency diffusion schedules, yielding higher sampling quality. Third, we show how this processes can be used to discard unwanted information at particular frequency ranges, yielding DPMs capable of extract noiseless signals from the remaining ranges. 

In our approach, we have limited the results to a simple two-band pass frequency filter. We propose in \autoref{supp:sec:freq_noise_operator} several other alternatives, which may serve as more flexible tools to inject useful inductive biases for similar tasks. Moreover, the approach can be extended beyond constant schedules. For instance, it may prove useful to introduce dynamic frequency noise strategies that shift the focus from low-frequency (general shapes) to high-frequency (sharp edges and textures) components over the time discretization of the sampling process. Such methods could more closely align with human visual processing, which progressively sharpens details over time, offering a more natural sampling process. Additionally, other domains of noise manipulation—outside of the frequency domain may also present new opportunities for further improving DPMs across various tasks.

Finally, a current limitation of this approach lies in the complexity of understanding the relationship between visual data in spatial and frequency domains. The perception of information in the frequency domain does not always translate straightforwardly to visual content, complicating the process of designing optimal noise schedules. As such, it is not trivial to design appropriate frequency schedules for a particular distribution. In practice, empirical validation may still be required to identify the best inductive biases for a given dataset. Future work could focus on refining analytical tools for frequency analysis or exploring alternative inductive bias mechanisms that extend beyond frequency-based manipulations.


Overall, this work opens the door for more targeted and flexible diffusion generative modeling by building inductive biases through the manipulation of the forward nosing process. The ability to design noise schedules that align with specific data characteristics holds promise for advancing the state of the art in generative modeling.


\subsubsection*{Acknowledgments}
The authors acknowledge funding from CIFAR, and Recursion. The research was enabled in part by computational resources provided by the Digital Research
Alliance of Canada (\url{https://alliancecan.ca}), Mila (\url{https://mila.quebec}), and
NVIDIA.

\bibliography{iclr2025_conference}
\bibliographystyle{iclr2025_conference}

\end{document}

%% file: math_commands.tex
\usepackage{amsmath,amsfonts,bm}

\def\eqref#1{equation~\ref{#1}}

\def\1{\bm{1}}

\def\rvx{{\mathbf{x}}}

\DeclareMathAlphabet{\mathsfit}{\encodingdefault}{\sfdefault}{m}{sl}
\SetMathAlphabet{\mathsfit}{bold}{\encodingdefault}{\sfdefault}{bx}{n}













%% file: iclr2025_workshop.bbl
\begin{thebibliography}{30}
\providecommand{\natexlab}[1]{#1}
\providecommand{\url}[1]{\texttt{#1}}
\expandafter\ifx\csname urlstyle\endcsname\relax
  \providecommand{\doi}[1]{doi: #1}\else
  \providecommand{\doi}{doi: \begingroup \urlstyle{rm}\Url}\fi

\bibitem[Alain \& Bengio(2014)Alain and Bengio]{alain2014regularized}
Guillaume Alain and Yoshua Bengio.
\newblock What regularized auto-encoders learn from the data-generating distribution.
\newblock \emph{The Journal of Machine Learning Research}, 15\penalty0 (1):\penalty0 3563--3593, 2014.

\bibitem[Bansal et~al.(2022)Bansal, Borgnia, Chu, Li, Kazemi, Huang, Goldblum, Geiping, and Goldstein]{bansal_cold_2022}
Arpit Bansal, Eitan Borgnia, Hong-Min Chu, Jie~S. Li, Hamid Kazemi, Furong Huang, Micah Goldblum, Jonas Geiping, and Tom Goldstein.
\newblock Cold {Diffusion}: {Inverting} {Arbitrary} {Image} {Transforms} {Without} {Noise}, August 2022.
\newblock URL \url{http://arxiv.org/abs/2208.09392}.
\newblock arXiv:2208.09392 [cs].

\bibitem[Bietti \& Mairal(2019)Bietti and Mairal]{bietti2019inductive}
Alberto Bietti and Julien Mairal.
\newblock On the inductive bias of neural tangent kernels.
\newblock In \emph{Advances in Neural Information Processing Systems}, 2019.

\bibitem[Crabbé et~al.(2024)Crabbé, Huynh, Stanczuk, and van~der Schaar]{crabbé2024timeseriesdiffusionfrequency}
Jonathan Crabbé, Nicolas Huynh, Jan Stanczuk, and Mihaela van~der Schaar.
\newblock Time series diffusion in the frequency domain, 2024.
\newblock URL \url{https://arxiv.org/abs/2402.05933}.

\bibitem[Geirhos et~al.(2019)Geirhos, Rubisch, Michaelis, Bethge, Wichmann, and Brendel]{geirhos2019imagenet}
Robert Geirhos, Patricia Rubisch, Claudio Michaelis, Matthias Bethge, Felix~A Wichmann, and Wieland Brendel.
\newblock Imagenet-trained cnns are biased towards texture; increasing shape bias improves accuracy and robustness.
\newblock In \emph{International Conference on Learning Representations}, 2019.

\bibitem[Geirhos et~al.(2020)Geirhos, Jacobsen, Michaelis, Zemel, Brendel, Bethge, and Wichmann]{Geirhos2020}
Robert Geirhos, J{\"o}rn-Henrik Jacobsen, Claudio Michaelis, Richard Zemel, Wieland Brendel, Matthias Bethge, and Felix~A. Wichmann.
\newblock Shortcut learning in deep neural networks.
\newblock \emph{Nature Machine Intelligence}, 2\penalty0 (11):\penalty0 665--673, 2020.
\newblock \doi{10.1038/s42256-020-00257-z}.
\newblock URL \url{https://doi.org/10.1038/s42256-020-00257-z}.

\bibitem[Ho et~al.(2020{\natexlab{a}})Ho, Jain, and Abbeel]{ho2020denoising}
Jonathan Ho, Ajay Jain, and Pieter Abbeel.
\newblock Denoising diffusion probabilistic models.
\newblock \emph{Advances in neural information processing systems}, 33:\penalty0 6840--6851, 2020{\natexlab{a}}.

\bibitem[Ho et~al.(2020{\natexlab{b}})Ho, Jain, and Abbeel]{ho_denoising_2020}
Jonathan Ho, Ajay Jain, and Pieter Abbeel.
\newblock Denoising {Diffusion} {Probabilistic} {Models}, June 2020{\natexlab{b}}.
\newblock URL \url{https://arxiv.org/abs/2006.11239v2}.

\bibitem[Kadkhodaie et~al.(2023)Kadkhodaie, Guth, Simoncelli, and Mallat]{kadkhodaie2023generalization}
Zahra Kadkhodaie, Florentin Guth, Eero~P Simoncelli, and St{\'e}phane Mallat.
\newblock Generalization in diffusion models arises from geometry-adaptive harmonic representation.
\newblock \emph{arXiv preprint arXiv:2310.02557}, 2023.

\bibitem[Krizhevsky et~al.(2009)Krizhevsky, Hinton, et~al.]{krizhevsky2009learning}
Alex Krizhevsky, Geoffrey Hinton, et~al.
\newblock Learning multiple layers of features from tiny images.
\newblock 2009.

\bibitem[Liu et~al.(2015)Liu, Luo, Wang, and Tang]{liu2015faceattributes}
Ziwei Liu, Ping Luo, Xiaogang Wang, and Xiaoou Tang.
\newblock Deep learning face attributes in the wild.
\newblock In \emph{Proceedings of International Conference on Computer Vision (ICCV)}, December 2015.

\bibitem[Matthey et~al.(2017)Matthey, Higgins, Hassabis, and Lerchner]{dsprites17}
Loic Matthey, Irina Higgins, Demis Hassabis, and Alexander Lerchner.
\newblock dsprites: Disentanglement testing sprites dataset.
\newblock https://github.com/deepmind/dsprites-dataset/, 2017.

\bibitem[Peng et~al.(2019)Peng, Bai, Xia, Huang, Saenko, and Wang]{peng2019moment}
Xingchao Peng, Qinxun Bai, Xide Xia, Zijun Huang, Kate Saenko, and Bo~Wang.
\newblock Moment matching for multi-source domain adaptation.
\newblock In \emph{Proceedings of the IEEE/CVF international conference on computer vision}, pp.\  1406--1415, 2019.

\bibitem[Qian et~al.(2024)Qian, Cai, Pan, Li, Yao, Sun, and Mei]{Qian_2024_CVPR}
Yurui Qian, Qi~Cai, Yingwei Pan, Yehao Li, Ting Yao, Qibin Sun, and Tao Mei.
\newblock Boosting diffusion models with moving average sampling in frequency domain.
\newblock In \emph{Proceedings of the IEEE/CVF Conference on Computer Vision and Pattern Recognition (CVPR)}, pp.\  8911--8920, June 2024.

\bibitem[Rahaman et~al.(2019)Rahaman, Baratin, Arpit, Draxler, Lin, Hamprecht, Bengio, and Courville]{rahaman_spectral_2019}
Nasim Rahaman, Aristide Baratin, Devansh Arpit, Felix Draxler, Min Lin, Fred Hamprecht, Yoshua Bengio, and Aaron Courville.
\newblock On the {Spectral} {Bias} of {Neural} {Networks}.
\newblock In \emph{Proceedings of the 36th {International} {Conference} on {Machine} {Learning}}, pp.\  5301--5310. PMLR, May 2019.
\newblock URL \url{https://proceedings.mlr.press/v97/rahaman19a.html}.
\newblock ISSN: 2640-3498.

\bibitem[Sahoo et~al.(2024)Sahoo, Gokaslan, De~Sa, and Kuleshov]{sahoo_diffusion_2024}
Subham~Sekhar Sahoo, Aaron Gokaslan, Chris De~Sa, and Volodymyr Kuleshov.
\newblock Diffusion {Models} {With} {Learned} {Adaptive} {Noise}, June 2024.
\newblock URL \url{http://arxiv.org/abs/2312.13236}.
\newblock arXiv:2312.13236 [cs].

\bibitem[Saleh \& Elgammal(2015)Saleh and Elgammal]{saleh2015large}
Babak Saleh and Ahmed Elgammal.
\newblock Large-scale classification of fine-art paintings: Learning the right metric on the right feature.
\newblock \emph{arXiv preprint arXiv:1505.00855}, 2015.

\bibitem[Scimeca et~al.(2021)Scimeca, Oh, Chun, Poli, and Yun]{scimeca2021which}
Luca Scimeca, Seong~Joon Oh, Sanghyuk Chun, Michael Poli, and Sangdoo Yun.
\newblock Which shortcut cues will dnns choose? a study from the parameter-space perspective.
\newblock In \emph{International Conference on Learning Representations}, 2021.

\bibitem[Scimeca et~al.(2023{\natexlab{a}})Scimeca, Rubinstein, Nicolicioiu, Teney, and Bengio]{scimeca2023leveraging}
Luca Scimeca, Alexander Rubinstein, Armand Nicolicioiu, Damien Teney, and Yoshua Bengio.
\newblock Leveraging diffusion disentangled representations to mitigate shortcuts in underspecified visual tasks.
\newblock In \emph{NeurIPS 2023 Workshop on Diffusion Models}, 2023{\natexlab{a}}.
\newblock URL \url{https://openreview.net/forum?id=AvUAVYRA70}.

\bibitem[Scimeca et~al.(2023{\natexlab{b}})Scimeca, Rubinstein, Teney, Oh, Nicolicioiu, and Bengio]{scimeca2023shortcut}
Luca Scimeca, Alexander Rubinstein, Damien Teney, Seong~Joon Oh, Armand~Mihai Nicolicioiu, and Yoshua Bengio.
\newblock Shortcut bias mitigation via ensemble diversity using diffusion probabilistic models.
\newblock \emph{arXiv preprint arXiv:2311.16176}, 2023{\natexlab{b}}.

\bibitem[Sendera et~al.(2024)Sendera, Kim, Mittal, Lemos, Scimeca, Rector-Brooks, Adam, Bengio, and Malkin]{sendera2024diffusion}
Marcin Sendera, Minsu Kim, Sarthak Mittal, Pablo Lemos, Luca Scimeca, Jarrid Rector-Brooks, Alexandre Adam, Yoshua Bengio, and Nikolay Malkin.
\newblock On diffusion models for amortized inference: Benchmarking and improving stochastic control and sampling.
\newblock \emph{arXiv preprint arXiv:2402.05098}, 2024.

\bibitem[Sohl-Dickstein et~al.(2015)Sohl-Dickstein, Weiss, Maheswaranathan, and Ganguli]{sohl-dickstein_deep_2015}
Jascha Sohl-Dickstein, Eric~A. Weiss, Niru Maheswaranathan, and Surya Ganguli.
\newblock Deep {Unsupervised} {Learning} using {Nonequilibrium} {Thermodynamics}, November 2015.
\newblock URL \url{http://arxiv.org/abs/1503.03585}.
\newblock arXiv:1503.03585 [cond-mat, q-bio, stat].

\bibitem[Tishby \& Zaslavsky(2015)Tishby and Zaslavsky]{tishby2015deep}
Naftali Tishby and Noga Zaslavsky.
\newblock Deep learning and the information bottleneck principle.
\newblock In \emph{2015 IEEE Information Theory Workshop (ITW)}, pp.\  1--5. IEEE, 2015.

\bibitem[Van~der Schaaf \& van Hateren(1996)Van~der Schaaf and van Hateren]{van1996modelling}
van~A Van~der Schaaf and JH~van van Hateren.
\newblock Modelling the power spectra of natural images: statistics and information.
\newblock \emph{Vision research}, 36\penalty0 (17):\penalty0 2759--2770, 1996.

\bibitem[Venkatraman et~al.(2024)Venkatraman, Jain, Scimeca, Kim, Sendera, Hasan, Rowe, Mittal, Lemos, Bengio, et~al.]{venkatraman2024amortizing}
Siddarth Venkatraman, Moksh Jain, Luca Scimeca, Minsu Kim, Marcin Sendera, Mohsin Hasan, Luke Rowe, Sarthak Mittal, Pablo Lemos, Emmanuel Bengio, et~al.
\newblock Amortizing intractable inference in diffusion models for vision, language, and control.
\newblock \emph{arXiv preprint arXiv:2405.20971}, 2024.

\bibitem[Vincent et~al.(2008)Vincent, Larochelle, Bengio, and Manzagol]{vincent2008extracting}
Pascal Vincent, Hugo Larochelle, Yoshua Bengio, and Pierre-Antoine Manzagol.
\newblock Extracting and composing robust features with denoising autoencoders.
\newblock In \emph{Proceedings of the 25th international conference on Machine learning}, pp.\  1096--1103, 2008.

\bibitem[Xu et~al.(2019)Xu, Zhang, and Xiao]{xu_training_2019}
Zhi-Qin~John Xu, Yaoyu Zhang, and Yanyang Xiao.
\newblock Training behavior of deep neural network in frequency domain, October 2019.
\newblock URL \url{http://arxiv.org/abs/1807.01251}.
\newblock arXiv:1807.01251 [cs, math, stat].

\bibitem[Yang et~al.(2024)Yang, Zhang, Song, Hong, Xu, Zhao, Zhang, Cui, and Yang]{yang_diffusion_2024}
Ling Yang, Zhilong Zhang, Yang Song, Shenda Hong, Runsheng Xu, Yue Zhao, Wentao Zhang, Bin Cui, and Ming-Hsuan Yang.
\newblock Diffusion {Models}: {A} {Comprehensive} {Survey} of {Methods} and {Applications}, June 2024.
\newblock URL \url{http://arxiv.org/abs/2209.00796}.
\newblock arXiv:2209.00796 [cs].

\bibitem[Yuan et~al.(2023)Yuan, Li, Wang, Yang, Lin, Liu, and Wang]{yuan2023spatialfrequencyunetdenoisingdiffusion}
Xin Yuan, Linjie Li, Jianfeng Wang, Zhengyuan Yang, Kevin Lin, Zicheng Liu, and Lijuan Wang.
\newblock Spatial-frequency u-net for denoising diffusion probabilistic models, 2023.
\newblock URL \url{https://arxiv.org/abs/2307.14648}.

\bibitem[Zhang et~al.(2024)Zhang, Zhang, Zhan, Luo, Wen, and Tao]{zhang2024confronting}
Ziyi Zhang, Sen Zhang, Yibing Zhan, Yong Luo, Yonggang Wen, and Dacheng Tao.
\newblock Confronting reward overoptimization for diffusion models: A perspective of inductive and primacy biases.
\newblock \emph{arXiv preprint arXiv:2402.08552}, 2024.

\end{thebibliography}
